\documentclass[10pt,twocolumn,letterpaper]{article}

\usepackage[pagenumbers]{cvpr}
\usepackage[accsupp]{axessibility}  

\usepackage{bm}
\usepackage{multirow}
\usepackage{makecell}
\usepackage{booktabs}

\definecolor{cvprblue}{rgb}{0.21,0.49,0.74}
\usepackage[pagebackref,breaklinks,colorlinks,allcolors=cvprblue]{hyperref}

\def\paperID{13423} 
\def\confName{CVPR}
\def\confYear{2026}

\title{CoVFT: Context-aware Visual Fine-tuning for Multimodal \\ Large Language Models}

\author{Nan Zhou$^{1,2}$\quad Huiqun Wang$^{1,2}$\quad Yaoyan Zheng$^{1,2}$\quad  Di Huang$^{1,2}$\thanks{Corresponding author.}\\
$^{1}$State Key Laboratory of Complex and Critical Software Environment, Beihang University \\
$^{2}$School of Computer Science and Engineering, Beihang University \\
{\tt\small \{zhounan0431,hqwangscse,yaoyanzheng,dhuang\}@buaa.edu.cn}
}

\begin{document}
\maketitle

\begin{abstract}
Multimodal large language models (MLLMs) achieve remarkable progress in cross-modal perception and reasoning, yet a fundamental question remains unresolved: should the vision encoder be fine-tuned or frozen? Despite the success of models such as LLaVA and Qwen-VL, inconsistent design choices and heterogeneous training setups hinder a unified understanding of visual fine-tuning (VFT) in MLLMs. Through a configuration-aligned benchmark, we find that existing VFT methods fail to consistently outperform the frozen baseline across multimodal tasks. Our analysis suggests that this instability arises from visual preference conflicts, where the context-agnostic nature of vision encoders induces divergent parameter updates under diverse multimodal context. To address this issue, we propose the \textbf{Context-aware Visual Fine-tuning (CoVFT)} framework, which explicitly incorporates multimodal context into visual adaptation. By integrating a Context Vector Extraction (CVE) and a Contextual Mixture-of-Experts (CoMoE) module, CoVFT decomposes conflicting optimization signals and enables stable, context-sensitive visual updates. Extensive experiments on 12 multimodal benchmarks demonstrate that CoVFT achieves state-of-the-art performance with superior stability. Notably, fine-tuning a 7B MLLM with CoVFT surpasses the average performance of its 13B counterpart, revealing substantial untapped potential in visual encoder optimization within MLLMs. 

Code: \url{https://github.com/weeknan/CoVFT}.

\end{abstract}  

\section{Introduction}
\label{sec:intro}

Recent advancements in multimodal large language models (MLLMs) demonstrate remarkable progress in cross-modal perception and reasoning \cite{qwen2.5vl, gemini2.5}. These models effectively integrate the visual perception capabilities of vision encoders with the reasoning and language understanding abilities of large language models (LLMs), yielding strong cross-modal comprehension and generation capabilities.

The current MLLM construction pipeline typically consists of two major stages \cite{mllm_survey}: a pre-training phase that aligns the vision encoder with the LLM, and an instruction-tuning phase that enhances cross-modal understanding and generation capabilities. A careful review of existing instruction-tuning practices reveals a notable divergence in visual fine-tuning~(VFT)--specifically, how to fine-tune the vision encoder. Building on pre-trained vision foundation models \cite{clip,siglip}, VFT can be regarded as adapting the vision encoder from pure visual perception to multimodal understanding in collaboration with the LLM. However, influential MLLMs adopt distinct VFT strategies. For instance, models such as InstructBLIP \cite{instructblip} and LLaVA-1.5 \cite{llava1.5} freeze the vision encoder during instruction-tuning, whereas InternVL \cite{internvl} and Qwen-VL \cite{qwen2.5vl} jointly optimize both the vision encoder and other components. This divergence highlights inconsistencies in adapting pre-trained vision encoder to MLLM applications, making the decision of whether to fine-tune or freeze the vision encoder a long-standing yet unresolved issue in the community.

\begin{figure*}[t]
  \centering 
  \includegraphics[width=172mm]{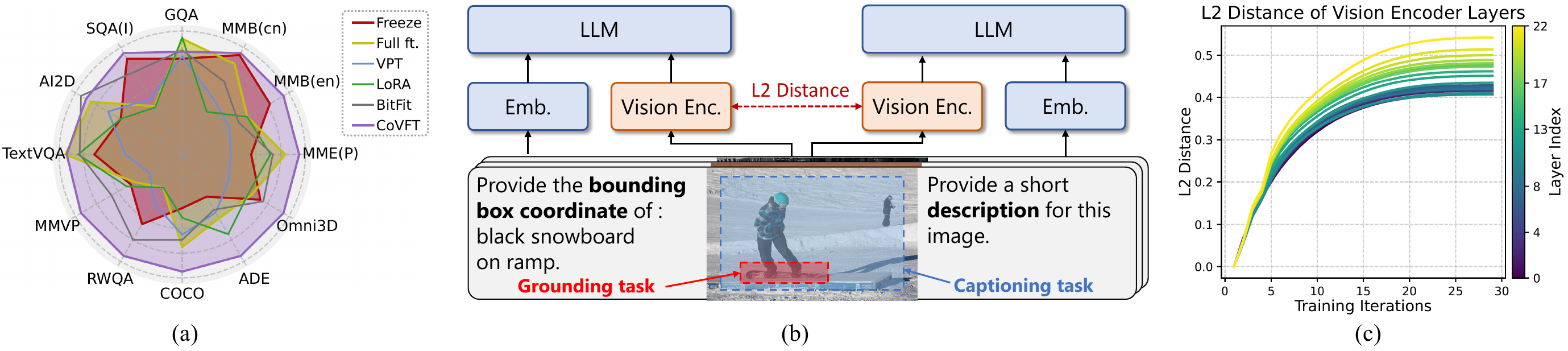}
  \caption{
  (a) Performance comparison of distinct VFT methods on multimodal tasks, showing instability compared with Freeze baseline.
  (b) Illustration of the \emph{visual preference conflict} phenomenon. Setup: we construct grounding and captioning tasks using identical images from the Visual Genome~\cite{Visualgenome} dataset and train two independent MLLMs that differ only in their textual queries. (c) The L2 distance between the two vision encoders steadily increases during VFT, indicating growing divergence in learned, especially the deeper representations, which aligns with the classical finding that deeper layers tend to capture information with task-specific preferences \cite{yosinski2014transferable, long2015learning}.
  } 
  \label{motivation}
\end{figure*}

To better understand this discrepancy, we follow previous studies \cite{cambrian,prismatic} and adopt the widely used LLaVA paradigm to systematically investigate the effect of different VFT strategies. Specifically, we compare three representative methods for training the vision encoder during instruction tuning: freezing, full fine-tuning, and parameter-efficient fine-tuning (PEFT) \cite{petl_survey}. Interestingly, our experiments in Fig.~\ref{motivation} (a) reveal that VFT methods rarely achieve consistent improvements over the frozen-encoder baseline in MLLM settings. This finding contradicts the well-established advantages of VFT observed in prior pure-vision transfer studies \cite{petl_survey_vision,vlm_survey}, where fine-tuning vision encoder consistently leads to better downstream performance, highlighting the absence of a stable and reliable VFT method for MLLMs.

Upon examining the update dynamics of the vision encoder during VFT, we observe key differences between VFT in MLLMs and pure vision scenarios. Visual–language tasks, which rely on linguistic supervision, are inherently more ambiguous than explicit vision-only tasks. In MLLMs, the VFT process heavily depends on contextual variation, as visual patterns differ significantly across diverse linguistic inputs. For example, grounding tasks \cite{Visualgenome} emphasize localizing fine-grained objects, whereas captioning tasks \cite{Mscococaptions} require capturing global semantics. Such diversity in context-driven visual preference is natural and desirable; however, when vision encoders are jointly fine-tuned across heterogeneous contexts, conflicting optimization signals emerge. These conflicts pull the encoder parameters in inconsistent directions, leading to unstable updates and degraded generalization. As illustrated in Fig.~\ref{motivation}(b) and (c), we term this phenomenon \emph{visual preference conflicts}.

Motivated by these observations, we aim to design a context-driven VFT framework that explicitly accounts for multimodal ambiguity and task-dependent variation. This requires addressing two key challenges. First, the vision encoder's updates should be guided not only by visual inputs but also by linguistic cues to achieve a more comprehensive understanding of the multimodal context. Second, the optimization process should remain generalizable across different types of visual preference conflicts and adaptable to varying perceptual patterns. To this end, we propose Context-aware Visual Fine-Tuning (CoVFT), a framework for context-adaptive visual adaptation in MLLMs. Specifically, we introduce a Contextual Vector Extraction (CVE) module that progressively derives a compact representation of multimodal context through text-guided cross-modal aggregation within the vision encoder. In parallel, a Contextual Mixture-of-Experts (CoMoE) module leverages this contextual signal to dynamically route visual tokens across multiple expert subspaces, effectively decomposing conflicting optimization signals into context-specialized updates. By integrating both modules, CoVFT aligns visual encoding with multimodal context, mitigating visual preference conflicts and enhancing multimodal capabilities.

Experiments on 12 multimodal benchmarks demonstrate that CoVFT achieves state-of-the-art and remarkably stable performance among existing VFT methods. Notably, fine-tuning the vision encoder of a 7B-scale MLLM with CoVFT surpasses the average performance of its 13B counterpart, underscoring the importance of vision encoder fine-tuning in MLLMs, even though the vision encoder accounts for less than 5\% of the total parameters. We hope this work sheds light on the underexplored yet essential problem of vision encoder training and inspires renewed attention to its design within the MLLM community.

\noindent The main contributions are summarized as follows:
\begin{itemize}[leftmargin=*]
    \item We systematically benchmark representative VFT methods under the LLaVA paradigm and identify the phenomenon of visual preference conflicts in MLLMs.
    \item We propose the CoVFT framework, which integrates the CVE and CoMoE modules to incorporate contextual information into visual adaptation, enabling adaptive and context-conditioned fine-tuning.
    \item We conduct experiments on 12 multimodal benchmarks, demonstrating that CoVFT not only outperforms state-of-the-art VFT methods but also exhibits strong data scalability and architectural generalizability.
\end{itemize}

\section{Related Work}

\subsection{Multimodal Large Language Models}

MLLMs integrate the visual perception capabilities of vision encoders with the reasoning and language understanding abilities of LLMs, thereby advancing cross-modal perception, understanding, and generation \cite{mllm_survey}. Researchers have made rapid progress in developing architectures that effectively fuse visual and linguistic representations. Early works such as Flamingo \cite{Flamingo} and the BLIP series \cite{blip, blip2, instructblip} employ cross-attention mechanisms and Q-Former structures to bridge visual and textual embeddings, enabling the LLM to condition its reasoning on image content. More recently, LLaVA and LLaVA-1.5 \cite{llava, llava1.5} simplify this design by replacing the Q-Former with a lightweight MLP projector, which substantially improves scalability and training stability. This paradigm has since been widely adopted due to its simplicity and competitive performance. Models such as Qwen-VL \cite{qwen3vl} and InternVL \cite{internvl} further strengthen vision-language alignment through multi-stage training pipelines and large-scale multimodal supervision, achieving significantly improved cross-modal reasoning capabilities.

Building on these advances, most current MLLMs are trained through a two-stage pipeline \cite{mllm_survey}. The pre-training stage leverages large-scale paired image–text data to align the projection layer and establish strong image–text correspondence. The subsequent instruction-tuning stage refines the LLM and projector with instruction-following data to enable higher-level reasoning and generalization. However, despite remarkable progress in large-scale pre-training and instruction-tuning, visual fine-tuning (VFT) strategies within MLLMs remain underexplored. InstructBLIP \cite{instructblip} and LLaVA \cite{llava,llava1.5} freeze the vision encoder, whereas InternVL \cite{internvl} and Qwen-VL \cite{qwen2.5vl} adopt end-to-end joint optimization. Consequently, existing approaches employ diverse and often inconsistent strategies, and prior studies report conflicting conclusions on whether freezing or fine-tuning the vision encoder yields better performance \cite{prismatic, cambrian}, leaving open the question of how best to adapt pre-trained vision encoders for multimodal reasoning.

\subsection{Visual Fine-tuning Methods}

In contrast to its limited exploration in MLLMs, VFT has been extensively studied in vision domains \cite{petl_survey_vision, hanfacing}. Mainstream VFT approaches can be broadly categorized into full parameter fine-tuning (Full FT) and linear probing (LP). Full FT updates all pre-trained model parameters, with subsequent works incorporating regularization techniques \cite{xuhong2018L2SP,li2018delta,distance_reg} to preserve pre-trained priors \cite{drtune}, or designing task-specific objectives \cite{zhan2018mix,gunel2020scl_tune} to enhance downstream adaptation. LP, by contrast, freezes the vision backbone and trains only a lightweight task-specific head. Inspired by LP, parameter-efficient fine-tuning (PEFT) methods introduce lightweight adaptation modules to balance performance and efficiency. These include partial tuning \cite{bitfit,spt}, which selectively updates sensitive parameters; prompt tuning \cite{jia2022vpt,e2vpt,svpt}, which learns visual prompts inserted into the input sequence; and adapter tuning \cite{adaptformer,lora}, which inserts additional low-rank or bottleneck layers into the network.

Despite the success of VFT in vision tasks \cite{hanfacing}, our configuration-aligned benchmark shows that existing VFT methods remain unstable in MLLMs. We attribute this instability to \emph{visual preference conflicts}, where a context-agnostic vision encoder receives conflicting update signals from diverse multimodal objectives. Although several works explore context-aware visual representations, such as conditioned visual embeddings for composed image retrieval \cite{composedimageretrieval} and text-conditioned queries for aggregating frozen visual features \cite{blip2, qmoe}, these methods are mainly developed for vision-language models (VLMs) \cite{clip} with limited reasoning capabilities or rely on frozen visual representations without adapting the vision encoder. Consequently, the impact of context-aware visual fine-tuning in large-scale MLLMs remains largely unexplored.

\begin{figure*}[t]
  \centering 
  \includegraphics[width=168mm]{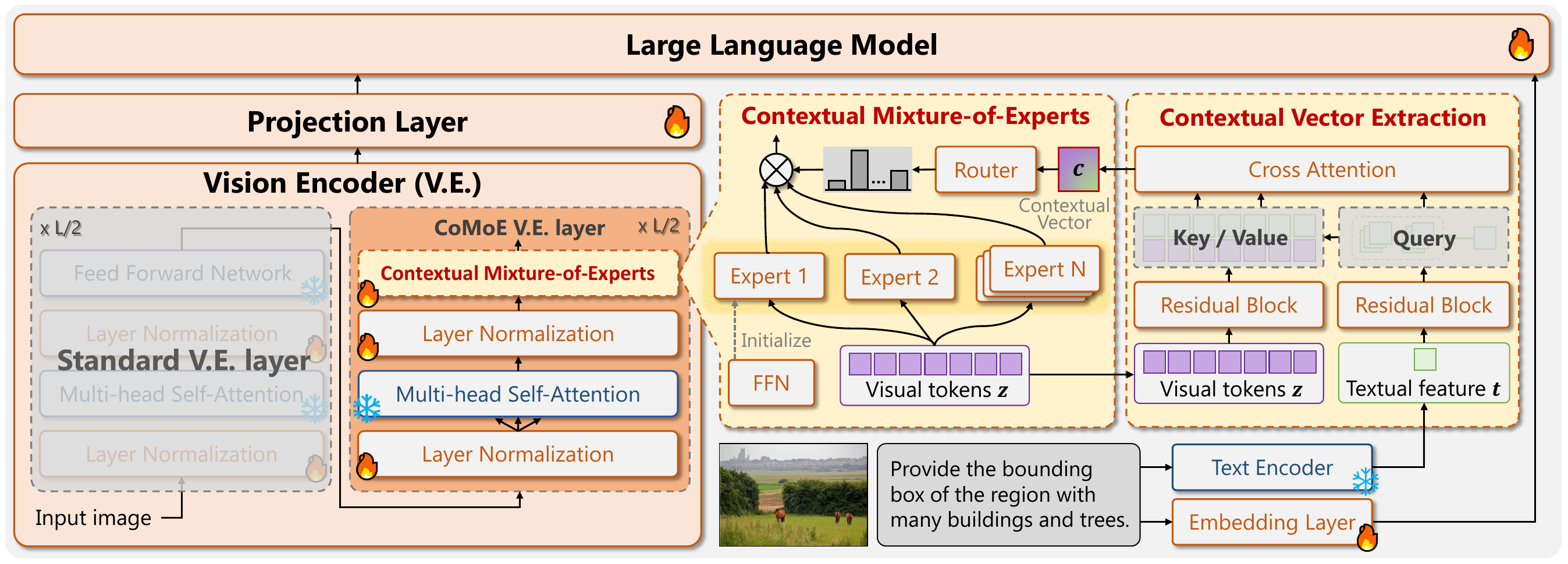}
  \caption{Illustration of the proposed context-aware visual fine-tuning (CoVFT) framework. Contextual vector extraction (CVE) generates a contextual vector $\bm{c}$ by aggregating multimodal cues through text-guided cross-attention. Contextual mixture-of-experts (CoMoE) injects $\bm{c}$ into the vision encoder via context-conditioned expert routing, enabling adaptive visual parameter updates.
  } 
  \label{fig:pipeline}
\end{figure*}

\section{Method}

\subsection{Formulation}

\label{sec:preliminaries}

\noindent\textbf{Architecture.} We consider an MLLM with LLaVA-style architecture \cite{llava1.5}, consisting of a vision encoder $\mathcal{V}$, a projector $\mathcal{P}$, and an LLM $\mathcal{L}$.  
Given an image $\mathbf{I}\!\in\!\mathbb{R}^{H\times W\times 3}$, the vision encoder extracts visual embeddings
\begin{equation}
    \bm{z} = \mathcal{V}(\mathbf{I}) \in \mathbb{R}^{L_v \times D_v},
\end{equation}
where $L_v$ and $D_v$ denote the sequence length and embedding dimension of the visual features, respectively. 
The projector $\mathcal{P}$ projects $\bm{z}$ into the embedding space of the LLM:
\begin{equation}
    \tilde{\bm{z}} = \mathcal{P}(\bm{z}) \in \mathbb{R}^{L_v \times D_t},
\end{equation}
where $D_t$ is the textual embedding dimension of the LLM.
The projected visual tokens are concatenated with the textual embeddings and jointly processed by the LLM:
\begin{equation}
    \mathbf{h} = \mathcal{L}([\tilde{\bm{z}}, {\rm Embed}(\text{text})]).
\end{equation}

\noindent\textbf{Training pipeline.} Most MLLMs are trained in two stages. The pre-training stage aligns the vision encoder with the LLM by training only the projector on paired image-caption data $(\mathbf{I}, \mathbf{C})$. The instruction-tuning stage further optimizes the MLLM on image-instruction-answer triplets $(\mathbf{I},\mathbf{Q},\mathbf{A})$, where $\mathbf{A}=\{a_1,\dots,a_T\}$ denotes the answer token sequence, with the next-token prediction loss:
\begin{equation}
\mathcal{L}_{\text{inst}} = - \sum_{t=1}^{T} \log p_{\bm \theta} (a_t \mid a_{<t}, \mathbf{Q}, \mathbf{I}),
\end{equation}
where ${\bm \theta} = \{\bm{\theta}_v, \bm{\theta}_p, \bm{\theta}_\ell\}$ are the parameters of the vision encoder, projector, and LLM, respectively.

\noindent\textbf{Visual fine-tuning.} 
In instruction-tuning, the vision encoder can either be frozen (\emph{i.e.} $\nabla_{\bm{\theta}_v} \mathcal{L}_{\text{inst}} = 0$) or fine-tuned jointly with other modules.  
Despite diverse practices, current VFT approaches all implicitly model a single posterior:
\begin{equation}
p_{\bm{\theta}_v}(\bm{z}\mid \mathbf{I}),
\end{equation}
and update $\bm{\theta}_v$ via $\nabla_{\bm{\theta}_v}\mathcal{L}_{\text{inst}}$.  
However, unlike pure vision fine-tuning, where supervision is derived solely from explicit visual labels, VFT in MLLMs depends on both image and linguistic context.  
Different instructions $\{\mathbf{q}_1, \mathbf{q}_2, \ldots\}$ induce distinct gradient directions $\nabla_{\bm{\theta}_v}^{(\mathbf{q}_i)}\mathcal{L}_{\text{inst}}$, guiding the encoder toward different visual cues or semantic regions and producing \emph{visual preference conflicts}.  
At an intuitive level, these conflicts correspond to inconsistent, task-dependent visual preferences; at the optimization level, they manifest as gradient misalignment that destabilizes VFT.

\subsection{Context-aware Visual Fine-tuning Framework}

To overcome the context-agnostic limitation of $p_{\bm{\theta}_v}(\bm{z}\mid\mathbf{I})$, we introduce a latent contextual variable $\bm{c}$ and reformulate the learning objective as a \emph{context-aware} posterior:
\begin{equation}
p_{\bm{\theta}_v}(\bm{z}\mid \mathbf{I},\bm{c}),
\end{equation}
where $\bm{c}$ encodes multimodal cues derived from the instruction triplet $(\mathbf{I}, \mathbf{Q}, \mathbf{A})$ and modulates visual representations accordingly. By conditioning the posterior on $\bm{c}$, the optimization of visual parameters becomes explicitly dependent on contextual semantics. This allows the vision encoder to adapt its feature updates according to the task-specific multimodal context, rather than relying on a single fixed mapping $p(\bm{z}\mid\mathbf{I})$, so that gradient directions are better aligned with the underlying visual-linguistic intent. Such context conditioning mitigates interference among heterogeneous multimodal tasks and leads to more stable adaptation.

\noindent\textbf{Framework overview.}  
As illustrated in Fig.~\ref{fig:pipeline}, the proposed Context-aware Visual Fine-tuning (CoVFT) framework instantiates this formulation through two complementary modules:  
(1)~Contextual Vector Extraction (CVE), which aggregates cross-modal information to produce $\bm{c}$ synchronously with the vision encoder, allowing text-guided attention to form a compact context vector that reflects task-specific visual preference;  
(2)~Contextual Mixture-of-Experts (CoMoE), which injects $\bm{c}$ into the encoder via context-conditioned expert routing. This design decomposes conflicting update directions into specialized expert subspaces, while encourages shared learning across experts with similar contextual patterns. Through this unified view, CoVFT models the context-aware posterior $p(\bm{z}\mid \mathbf{I},\bm{c})$, providing a principled framework for stable and adaptive visual encoder optimization in MLLMs.

\subsection{Contextual Vector Extraction}
\label{sec:cve}

The goal of contextual vector extraction (CVE) is to model the latent variable $\bm{c}$ as an explicit contextual vector for guiding context-aware visual adaptation.  
A desirable contextual vector should satisfy two properties:  
(1) it should evolve synchronously with the vision encoder during forward propagation, avoiding additional reasoning stages;  
(2) it should encode multimodal context, especially task-specific cues implied by the textual instruction, rather than being dominated by raw visual features.  
Given a textual instruction $\mathbf{q}$, a frozen text encoder (e.g., BERT~\cite{bert}) produces the textual embedding $\bm{t}$.  
At a certain layer of the vision encoder, visual tokens $\bm{z}$ and textual embedding $\bm{t}$ are first refined by a lightweight residual block:
\begin{equation}
    f_{\text{res}}(\mathbf{x}) = \mathbf{x} + {\rm GELU}(\mathbf{x} \cdot \mathbf{W}_{\rm up}) \mathbf{W}_{\rm down},
\end{equation}
yielding $\hat{\bm{z}}$ and $\hat{\bm{t}}$. A cross-attention operation then extracts the contextual vector by using $\hat{\bm{t}}$ as the query and the concatenated multimodal features as keys and values:
\begin{equation}
    \bm{c}_i = {\rm CrossAttn}\big({\rm norm}(\hat{\bm{t}})_{\rm q}, \; [{\rm norm}(\hat{\bm{z}}),{\rm norm}(\hat{\bm{t}})]_{\rm k,v}\big),
\end{equation}
where ${\rm norm}(\cdot)$ denotes LayerNorm~\cite{layernorm} and $[\cdot, \cdot]$ indicates channel-wise concatenation. The resulting $\bm{c}$ is updated layer by layer as the encoder processes the image, ensuring that contextual information remains tightly coupled with the visual inference flow. As shown in Fig.~\ref{fig:analysis_fig} (a), the extracted contextual vectors exhibit coherent semantic clustering and align well with task-specific visual preference patterns, indicating that $\bm{c}$ provides an effective guidance signal for expert routing in CoMoE.
 
\subsection{Contextual Mixture-of-Experts}
\label{sec:comoe}

While the contextual vector $\bm{c}$ provides multimodal cues, the original ViT \cite{vit} lacks mechanisms to incorporate non-visual conditioning into its parameter updates. We introduce the contextual mixture-of-experts (CoMoE) module to integrate $\bm{c}$ into the vision encoder, enabling context-aware feature extraction. As observed in Fig.~\ref{motivation} (c), visual preference conflicts are more pronounced in deeper layers. Therefore, starting from a designated layer, we replace the FFN with a set of $N$ parallel expert networks $\{\mathcal{E}^1,\dots,\mathcal{E}^N\}$, each initialized from the original FFN. Given $\bm{c}$, a learnable router computes context-conditioned routing weights:
\begin{equation}
    \bm{g}(\bm{c}) = {\rm softmax}(\mathbf{W}\bm{c} + \bm{b}) \in \mathbb{R}^{N}.
\end{equation}
Each expert processes the visual tokens $\bm{z}$, and their outputs are combined through dense aggregation:
\begin{equation}
    \tilde{\bm{z}} = \sum_{n=1}^{N} g^{n}(\bm{c})\,\mathcal{E}^n(\bm{z}).
\end{equation}
We empirically find that the above dense routing performs better than the more commonly used sparse routing (e.g., top-$k$ activation~\cite{switch}). A possible reason is that sparse routing reduces the number of samples assigned to each expert and may therefore lead to under-training when data per expert is limited, whereas dense routing allows all experts to receive gradients from every sample. Crucially, by computing routing weights from the contextual vector $\bm{c}$, CoMoE modulates the gradients received by each expert in a context-dependent manner. Specifically, for the $n$-th expert with parameters $\bm{\theta}_e^{n}$, its gradient under context $\bm{c}$ is scaled by the corresponding routing weight:
\begin{equation}
\nabla_{\bm{\theta}_e^{n}} \mathcal{L}_{\text{inst}}
=
g^{n}(\bm{c})\,\cdot
\frac{\partial \mathcal{L}_{\text{inst}}}{\partial \tilde{\bm{z}}}
\frac{\partial \mathcal{E}^{n}(\bm{z})}{\partial \bm{\theta}_e^{n}}.
\end{equation}
Thus, samples with similar contexts yield similar routing patterns and contribute more consistently to the same expert subspaces, while samples with different contexts are separated through differentiated gradient scaling. This context-aware gradient modulation alleviates interference among conflicting visual preferences, yielding more coherent convergence and more stable visual fine-tuning.

\section{Experiments}

\begin{table*}[t]\footnotesize
	\centering 
  \setlength{\tabcolsep}{2.7pt}
  \caption{
  Comparison of diverse VFT methods on 12 multimodal benchmarks under the unified LLaVA-1.5 ~\cite{llava1.5} pre-training and instruction-tuning pipeline. The best and second-best VFT results are highlighted in \colorbox{gray!20}{\textbf{bold}} and \underline{underline}, respectively. ``G.'', ``K.'', ``V.'', and ``Avg.'' denote the mean scores of the three task groups, and the overall average. Results marked with ``$\dagger$'' are obtained from LLaVA-1.5 or evaluating its official checkpoints, while ``$*$'' indicates our reproduction of LLaVA-1.5 Freeze baseline under the same configuration.
  }
	\begin{tabular}{l c c c c c c c c c c c c c c c c} 
      \toprule     
      ~ & \multicolumn{4}{c}{\textbf{General}} & \multicolumn{3}{c}{\textbf{Knowledge \& OCR}} & \multicolumn{5}{c}{ \textbf{Vision-centric}} & \multicolumn{4}{c}{\textbf{Mean (\%)}} \\
      \cmidrule(lr{0pt}){2-5} \cmidrule(lr{0pt}){6-8} \cmidrule(lr{0pt}){9-13} \cmidrule(lr{0pt}){14-17}
      Method & MME$^{\text P}$ & MMB$^\text{en}$ & MMB$^\text{cn}$ & GQA & SQA$^\text{I}$ & AI2D & TextVQA & MMVP & RWQA & COCO & ADE & Omni3D & \textbf{G.} & \textbf{K.} & \textbf{V.} & \textbf{Avg.} \\ 
      \midrule
      \makecell[l]{\textcolor{gray}{LLaVA-1.5-7B$^\dagger$}} & \textcolor{gray}{1510.7} & \textcolor{gray}{64.30} & \textcolor{gray}{58.30} & \textcolor{gray}{62.00} & \textcolor{gray}{66.80} & \textcolor{gray}{55.21} & \textcolor{gray}{58.20} & \textcolor{gray}{27.33} & \textcolor{gray}{56.21} & \textcolor{gray}{67.45} & \textcolor{gray}{54.98} & \textcolor{gray}{63.25}& \textcolor{gray}{65.04}& \textcolor{gray}{60.07} & \textcolor{gray}{53.84} & \textcolor{gray}{59.13}\\
      \makecell[l]{Freeze$^*$} & 1473.7 & \underline{67.87} & \underline{60.30} & 63.07 & \underline{69.31} & 55.76 & 58.53 & 28.00 & {56.73} & 63.73 & 49.61 & 60.50 & 66.23 & 61.20 & 51.71 & 58.93\\
      \makecell[l]{Full fine-tuning} & \underline{1510.2} & 67.44 & 59.88 & \underline{63.92} & 67.67 & 56.51 & \cellcolor{gray!20}\textbf{59.70} & 27.33 & 55.82 & \underline{65.71} & 51.97 & 60.00 & \underline{66.69} & 61.29 & 52.17 & 59.29\\
      \makecell[l]{VPT} & 1452.1 & 67.10 & 58.16 & 63.17 & 67.92 & 56.08 & 57.29 & 24.67 & 56.21 & 65.09 & 51.34 & 58.67 & 65.26 & 60.43 & 51.20 & 58.19\\
      \makecell[l]{SVPT} & 1431.7 & 66.58 & 57.90 & 63.26 & 68.82 & 56.54 & 57.89 & 29.33 & {56.86} & 65.34 & 51.03 & 59.29 & 64.83 & 61.08 & 52.37 & 58.70\\
      \makecell[l]{LoRA} & 1494.5 & 67.44 & 57.56 & \cellcolor{gray!20}\textbf{63.98} & 67.63 & 55.76 & 59.18 & 28.66 & 55.82 & 64.22 & \underline{53.71} & 59.83 & 65.93 & 60.86 & 52.45 & 59.04\\
      \makecell[l]{BitFit} & 1496.8 & 67.27 & 59.02 & 63.43 & 68.86 & \cellcolor{gray!20}\textbf{56.77} & 59.12 & \underline{31.33} & \underline{57.12} & 65.34 & 51.03 & \underline{60.67} & 66.14 & \underline{61.58} & \underline{53.10} & \underline{59.57}\\
      \makecell[l]{CoVFT} & \cellcolor{gray!20}\textbf{1525.2} & \cellcolor{gray!20}\textbf{68.13} & \cellcolor{gray!20}\textbf{60.40} & 63.37 & \cellcolor{gray!20}\textbf{69.51} & \underline{56.64} & \underline{59.64} & \cellcolor{gray!20}\textbf{36.67} & \cellcolor{gray!20}\textbf{57.52} & \cellcolor{gray!20}\textbf{66.96} & \cellcolor{gray!20}\textbf{56.08} & \cellcolor{gray!20}\textbf{61.83} & \cellcolor{gray!20}\textbf{67.04} & \cellcolor{gray!20}\textbf{61.93} & \cellcolor{gray!20}\textbf{55.81} & \cellcolor{gray!20}\textbf{61.08}\\
      \midrule
      \makecell[l]{\textcolor{gray}{LLaVA-1.5-13B$^\dagger$}} & \textcolor{gray}{1531.3} & \textcolor{gray}{67.70} & \textcolor{gray}{63.60} & \textcolor{gray}{63.30} & \textcolor{gray}{71.60} & \textcolor{gray}{59.16} & \textcolor{gray}{63.10} & \textcolor{gray}{30.67} & \textcolor{gray}{55.29} & \textcolor{gray}{66.82} & \textcolor{gray}{47.95} & \textcolor{gray}{63.00}& \textcolor{gray}{67.79}& \textcolor{gray}{64.62} & \textcolor{gray}{52.75} & \textcolor{gray}{60.73}\\
      \makecell[l]{Freeze$^*$} & \underline{1557.6} & 69.93 & \underline{63.57} & {63.73} & \underline{72.43} & \underline{60.20} & 60.18 & 34.00 & \underline{55.94} & \underline{65.09} & \cellcolor{gray!20}\textbf{51.34} & 62.83 & \underline{68.78} & 64.27 & 53.84  & \underline{61.43} \\
      \makecell[l]{Full fine-tuning} & 1555.4 & 68.56 & 63.49 & \cellcolor{gray!20}\textbf{64.83} & 71.99 & {59.94} & \underline{61.36} & 34.00 & 55.82 & 64.22 & \underline{50.39} & 63.25 & 68.66 & \underline{64.43} & {53.54}  & 61.30 \\
      \makecell[l]{BitFit} & 1552.8 & \underline{70.02} & 62.54 & \underline{63.98} & 72.14 & {59.52} & 61.04 & \underline{38.00} & 54.64 & 63.22 & 47.87 & \underline{66.50} & 68.55 & {64.23} & \underline{54.05} & \underline{61.43}\\
      \makecell[l]{CoVFT} & \cellcolor{gray!20}\textbf{1590.6} & \cellcolor{gray!20}\textbf{70.45} & \cellcolor{gray!20}\textbf{64.52} & 63.84 & \cellcolor{gray!20}\textbf{72.90} & \cellcolor{gray!20}\textbf{60.65} & \cellcolor{gray!20}\textbf{61.79} & \cellcolor{gray!20}\textbf{38.67} & \cellcolor{gray!20}\textbf{58.04} & \cellcolor{gray!20}\textbf{65.22} & \cellcolor{gray!20}\textbf{51.34} & \cellcolor{gray!20}\textbf{67.58} & \cellcolor{gray!20}\textbf{69.66} & \cellcolor{gray!20}\textbf{65.11} & \cellcolor{gray!20}\textbf{56.17} & \cellcolor{gray!20}\textbf{62.90}\\
      \bottomrule 
    \end{tabular}
  \label{all_compare}
\end{table*}

\vspace{-2pt}
\subsection{Experimental Setup}
\label{:sec:exp_setup}

\noindent \textbf{Visual fine-tuning benchmark.} To provide a comprehensive evaluation of representative VFT methods in MLLMs, we adopt the standard training configuration of LLaVA-1.5 \cite{llava1.5} and consider the following approaches: (1) Freeze, the default strategy used in LLaVA-1.5; (2) Full fine-tuning; and (3) PEFT methods covering three major paradigms, including partial tuning (Bitfit \cite{bitfit}), prompt-based tuning (VPT \cite{jia2022vpt} and SVPT \cite{svpt}, the latter reports state-of-the-art PEFT results on image classification \cite{vtab}), and adapter tuning (LoRA \cite{lora}). We employ CLIP-ViT-L/14-336 \cite{clip} as the vision encoder, a two-layer MLP as the projection layer, and Vicuna-1.5 \cite{vicuna} as the LLM.

Evaluation includes 12 multimodal benchmarks spanning three domains: (1) \textbf{General VQA:} MME-Perception (MME$^{\rm P}$) \cite{mme}, MMbench (MMB) \cite{mmbench}, and GQA \cite{gqa}; (2) \textbf{Knowledge and OCR:} ScienceQA-Image (SQA$^{\rm I}$) \cite{sqa}, AI2D \cite{ai2d}, and TextVQA \cite{textvqa}; (3) \textbf{Vision-centric tasks:} MMVP \cite{mmvp}, RealWorldQA (RWQA) \cite{realworldqa}, and CV-Bench (COCO, ADE, and Omni3D) \cite{cambrian}. 


\noindent \textbf{Training details.} Pre-training uses 558K image–caption pairs, where only the projector is optimized with a learning rate of $1 \times 10^{-3}$ and a batch size of 256. Instruction-tuning uses 665K image–instruction–answer triplets, during which the LLM and projector are jointly trained with a learning rate of $2 \times 10^{-5}$ and a batch size of 128, while the vision encoder is updated according to each VFT method using its default hyperparameters for fairness.

As for CoVFT, we apply the proposed CoMoE modules to the latter half of the ViT blocks by default. Each CoMoE layer comprises four expert networks. For the CVE module, we use the $\texttt{[CLS]}$ token from a frozen BERT text encoder \cite{bert} (though other text encoders are also compatible) to derive the textual representation. During instruction tuning, the CVE and CoMoE modules, together with the LayerNorm statistics of the vision encoder, are optimized jointly, while all other vision parameters remain frozen.

\subsection{Main Results}

The detailed comparisons are presented in Table~\ref{all_compare}. From these results, we make three key observations:

(1)~\textbf{Superiority and instability of VFT.} Although several works choose to freeze the vision encoder, our benchmark shows that most VFT methods—including full fine-tuning, LoRA, and BitFit—outperform the default Freeze strategy in average performance. This trend aligns with the findings of Cambrian-1~\cite{cambrian} and suggests that VFT is generally beneficial for modular MLLMs, as the pre-training objective of the vision encoder often diverges from the multimodal instruction-following objective. Without visual fine-tuning, the vision encoder functions as a static bottleneck, forcing the LLM to bear the full burden of refining visual representations. However, the task-level results reveal significant instability: full fine-tuning, LoRA, and BitFit surpass the Freeze baseline on only 6, 6, and 9 out of 12 benchmarks, respectively. This inconsistency contrasts sharply with the stable benefits of fine-tuning observed in pure vision domains~\cite{drtune,propetl}, indicating that existing VFT methods share a common limitation in overlooking context-dependent supervision and thus struggle to maintain stable cross-task generalization.

(2)~\textbf{Consistent and leading performance of CoVFT.} Compared with the Freeze baseline, CoVFT consistently achieves improvements across all 12 benchmarks, yielding an average gain of 2.15\%. Among all VFT methods, CoVFT attains state-of-the-art performance on 9 out of 12 tasks and all three task groups, achieving the highest overall average score of 61.08\%. These results demonstrate that context-aware visual fine-tuning effectively mitigates visual preference conflicts and enhances the adaptability of the vision encoder to diverse multimodal contexts.

(3)~\textbf{Scaling and capacity benefits of CoVFT.} Notably, fine-tuning the vision encoder of a 7B-scale MLLM with CoVFT surpasses the average performance of its 13B counterpart under the Freeze setting. This observation highlights the substantial untapped potential of VFT: the vision encoder accounts for less than 5\% of the total parameters in the 7B model, yet its optimization brings remarkable performance gains without scaling language parameters. We further evaluate CoVFT on the 13B model and compare it with both the Freeze baseline, Full fine-tuning and BitFit—the best-performing VFT method on the 7B model. CoVFT consistently achieves superior results over both, confirming that context-aware visual adaptation remains effective even at larger scales. These findings suggest that optimizing vision encoders is an underexplored yet promising direction for advancing MLLMs.

\begin{table}[t]\small  
    \centering 
    \caption{Ablation study of the contextual vector design in CVE and the routing strategy in CoMoE under LLaVA-1.5-7B.}
    \setlength{\tabcolsep}{5.5pt}
    \begin{tabular}{l c c c c} 	
        \toprule
        Ablation & General & {Know.\&OCR} & Vision & Mean (\%)\\
        \midrule
        N/A & 66.69 & 61.29 & 52.17 & 59.29 \\
        \multicolumn{5}{l}{\textcolor{gray}{\emph{w.r.t.} contextual vector}}\\   
        Image-only & 66.60 & 61.69 & 53.17 & 59.77 \\
        Text-only & 66.84 & 61.86 & 54.73 & 60.55 \\
        Concat$[\rm{I,T}]$ & 66.78 & 61.79 & 54.56 & 60.44 \\
        CVE & \textbf{67.04} & \textbf{61.93} & \textbf{55.81} & \textbf{61.08} \\
        \midrule
        \multicolumn{5}{l}{\textcolor{gray}{\emph{w.r.t.} routing strategy}}\\
        Random@2 & 66.01 & 61.24 & 52.05 & 59.00 \\
        Uniform & 66.18 & 61.75 & 53.05 & 59.60 \\
        Sparse@2 & 66.63 & 61.78 & 53.60 & 60.10 \\
        Dense & \textbf{67.04} & \textbf{61.93} & \textbf{55.81} & \textbf{61.08} \\ 
        \bottomrule 
    \end{tabular}
    \vspace{-4pt}
    \label{tab:design_choice}
\end{table}

\begin{figure*}[t]
  \centering 
  \includegraphics[width=170mm]{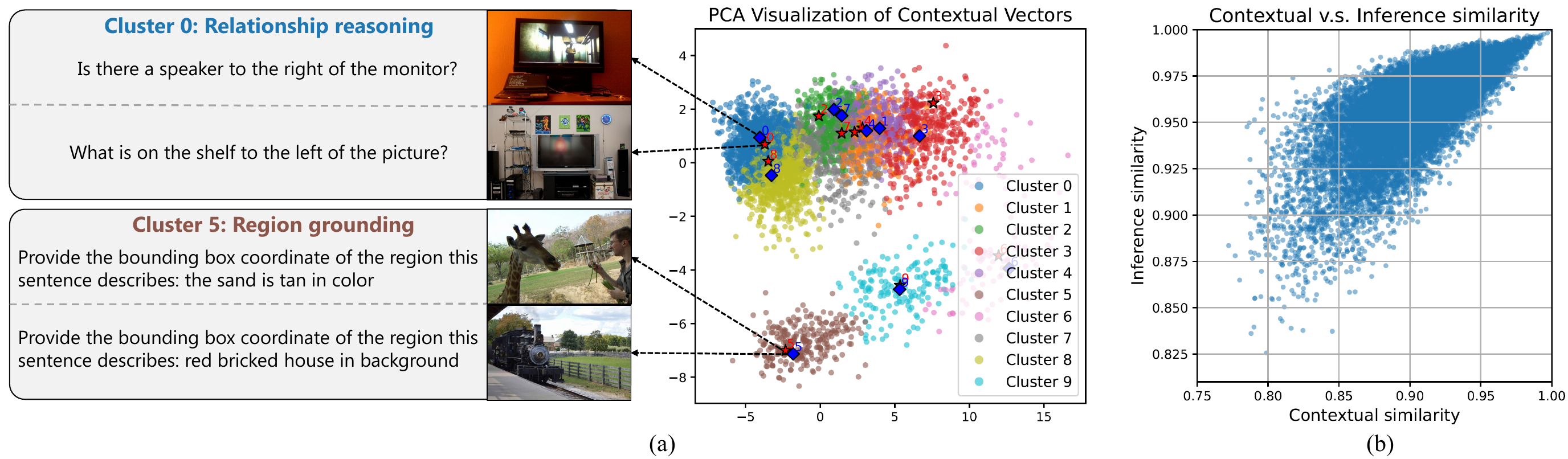}
  \caption{(a) PCA \cite{pca} visualization of contextual vectors extracted from the CVE module. A subset of 5,000 instruction samples is clustered via \emph{k}-means \cite{kmeans}, showing clear semantic grouping aligned with distinct visual preference patterns. (b) Correlation between contextual similarity and inference similarity, computed as the cosine similarity of routing weights aggregated across CoMoE layers. The strong positive trend ($r=0.76$) indicates that samples with similar contextual vectors yield similar expert activation patterns.
}
\label{fig:analysis_fig}
  
\end{figure*}

\subsection{Ablation Study}

To investigate the impact of the key design choices in CoVFT, we conduct ablation studies focusing on: (1) the formulation of the contextual vector $\bm{c}$ in the CVE module, and (2) the routing strategy in the CoMoE module.

\noindent \textbf{Contextual vector design.}
We examine several strategies for constructing the contextual vector $\bm{c}$, including averaging visual tokens (\emph{Image-only}), using the text encoder output (\emph{Text-only}), directly concatenating image and text features (\emph{Concat}), and the proposed text-guided aggregation in \emph{CVE}. The baseline (\emph{N/A}) corresponds to full fine-tuning the vision encoder without auxiliary context. As shown in Table~\ref{tab:design_choice}, incorporating textual features yields clear gains over the \emph{Image-only} variant, indicating that linguistic cues provide complementary context that mitigates visual preference conflicts. The naive concatenation of multimodal features (\emph{Concat}) performs comparably to \emph{Text-only}, suggesting that simple feature fusion cannot effectively capture fine-grained cross-modal dependencies. The proposed \emph{CVE} achieves the highest overall performance and shows notable improvements on vision-centric benchmarks, confirming that text-guided visual aggregation is essential for generating context-aware representations and fully exploiting visual information in multimodal settings.

\noindent \textbf{Routing strategy in CoMoE.}
We compare different routing mechanisms within CoMoE, including sparse activation of the top-2 experts (\emph{Sparse@2}) and dense activation (\emph{Dense}) of all experts weighted by routing weights. We also evaluate variants that equally averages all expert outputs (\emph{Uniform}) or randomly activates two experts (\emph{Random@2}) to isolate the effect of the additional parameters introduced by the MoE architecture. As shown in Table~\ref{tab:design_choice}, \emph{Dense} consistently outperforms \emph{Sparse@2}, suggesting that activating all experts enables better multimodal understanding and alleviates the under-training issue~\cite{xft}. Despite this, we note that \emph{Sparse@2} still performs better than existing VFT methods in Table~\ref{all_compare}, implying that sparse routing remains viable. Exploring larger-scale data to balance performance with the efficiency advantages of sparse inference presents a promising direction for future work. In contrast, \emph{Random@2} and \emph{Uniform}, despite increasing the model’s width, fail to yield stable gains, indicating that merely expanding the parameter space cannot resolve visual preference conflicts.

\begin{figure*}[t]
  \centering 
  \includegraphics[width=172mm]{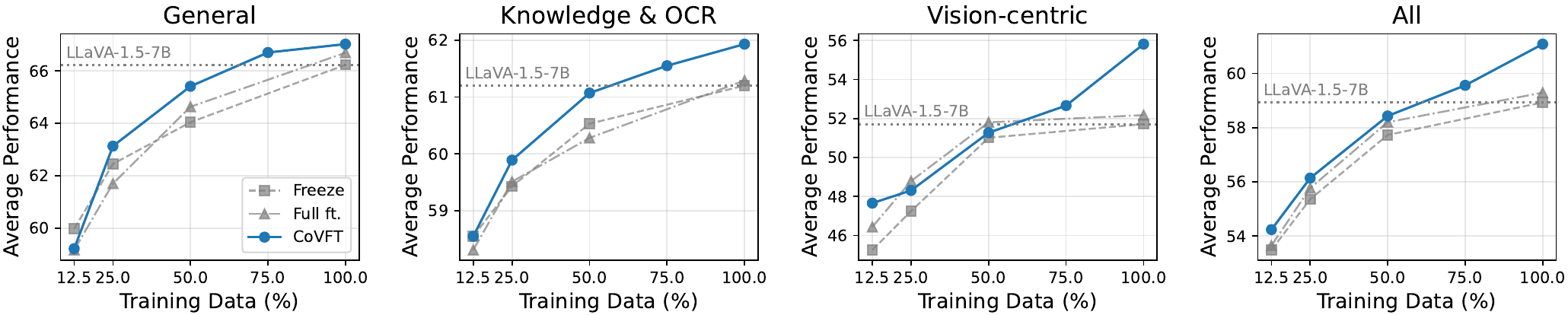}
  \caption{Analysis of data scalability. Following the two-stage LLaVA training pipeline, models are fine-tuned with varying proportions of the 665K multimodal instruction dataset during the second-stage instruction-tuning. We compare the performance of the baseline model with a frozen or full fine-tuned vision encoder and our CoVFT framework under different data scales.} 
  \label{fig:diff_data}
\end{figure*}

\begin{table*}[t]\footnotesize
	\centering 
  \setlength{\tabcolsep}{2.1pt}
  \caption{Generalization evaluation of CoVFT across different MLLM architectures on 12 multimodal benchmarks.
  }
	\begin{tabular}{l c c c c c c c c c c c c c c c c c} 
      \toprule     
       MLLM & {V. E. \& LLM} & VFT &  MME$^{\text P}$ & MMB$^\text{en}$ & MMB$^\text{cn}$ & GQA & SQA$^\text{I}$ & AI2D & TextVQA & MMVP & RWQA & COCO & ADE & Omni3D & Mean (\%) \\ 
      \midrule
       \multirow{2}{*}{LLaVA-1.5} & SigLIP-L/16 & Freeze & 1393.4 & {65.81} & {58.33} & 62.32 & \cellcolor{gray!20}\textbf{70.85} & 55.67 & 58.79 & 28.00 & {56.60} & 61.37 & 46.45 & 56.67 & 57.54\\
      & Vicuna-7B & CoVFT &  \cellcolor{gray!20}\textbf{1466.8} & \cellcolor{gray!20}\textbf{66.92} & \cellcolor{gray!20}\textbf{60.74} & \cellcolor{gray!20}\textbf{62.43} & 69.96 & \cellcolor{gray!20}\textbf{55.73} & \cellcolor{gray!20}\textbf{61.57} & \cellcolor{gray!20}\textbf{35.33} & \cellcolor{gray!20}\textbf{57.65} & \cellcolor{gray!20}\textbf{65.71} & \cellcolor{gray!20}\textbf{50.71} & \cellcolor{gray!20}\textbf{56.83} & \cellcolor{gray!20}\textbf{59.74}\\
      \midrule
      \multirow{2}{*}{LLaVA-1.5} & DINOv3-L/16 & Freeze & 1237.4 & 57.56 & 51.03 & 60.43 & \cellcolor{gray!20}\textbf{65.20} & \cellcolor{gray!20}\textbf{53.76} & 47.22 & 15.33 & {47.58} & 58.26 & 49.45 & 59.25  & 52.25\\
        & Vicuna-7B& CoVFT & \cellcolor{gray!20}\textbf{1309.8} & \cellcolor{gray!20}\textbf{58.33} & \cellcolor{gray!20}\textbf{51.80} & \cellcolor{gray!20}\textbf{62.50} & {64.90} & {52.95} & \cellcolor{gray!20}\textbf{49.75} & \cellcolor{gray!20}\textbf{20.00} & \cellcolor{gray!20}\textbf{52.55} & \cellcolor{gray!20}\textbf{67.08} & \cellcolor{gray!20}\textbf{50.08} & \cellcolor{gray!20}\textbf{62.17} & \cellcolor{gray!20}\textbf{54.80}\\
      \midrule
       \multirow{2}{*}{InternVL 2.0} & InternViT-300M & Full ft.& 1502.2 & {71.99} & 66.75 & 63.65 & {90.18} & {75.39} & {67.51} & 37.33 & {60.78} & {72.05} & {58.93} & 70.42 & 67.51 \\
         & Phi-3-mini & CoVFT&  \cellcolor{gray!20}\textbf{1527.0} & \cellcolor{gray!20}\textbf{75.95} & \cellcolor{gray!20}\textbf{69.33} & \cellcolor{gray!20}\textbf{64.47} & \cellcolor{gray!20}\textbf{94.25} & \cellcolor{gray!20}\textbf{77.46} & \cellcolor{gray!20}\textbf{71.16} & \cellcolor{gray!20}\textbf{44.00} & \cellcolor{gray!20}\textbf{62.35} & \cellcolor{gray!20}\textbf{72.80} & \cellcolor{gray!20}\textbf{60.35} & \cellcolor{gray!20}\textbf{72.83} & \cellcolor{gray!20}\textbf{70.11}\\
      \bottomrule 
    \end{tabular}
  \label{model_generalize}
\end{table*}

\subsection{Analysis}
\label{sec:analysis}

\noindent \textbf{Contextual vector.} To assess whether the contextual vector captures task-dependent multimodal cues and effectively guides expert routing in CoMoE, we conduct a two-stage analysis. As shown in Fig.~\ref{fig:analysis_fig} (a), the contextual vectors extracted from the CVE module form clear clusters, where samples within the same cluster tend to share similar visual preference requirements. The intra-cluster cosine similarity of visual and textual pairs increases by 19.23\% and 10.67\%, respectively, relative to inter-cluster pairs, indicating that the contextual vector encodes coherent multimodal semantics. Qualitative inspection further reveals that different clusters correspond to distinct visual preference patterns, such as relationship reasoning (requiring structural understanding) and region grounding (focusing on spatial localization). These observations suggest that the contextual vector captures task-dependent visual preferences and helps organize the heterogeneous multimodal space into more semantically coherent subregions.

Moreover, as shown in Fig.~\ref{fig:analysis_fig} (b), inference similarity exhibits a positive correlation with contextual similarity, indicating that samples with similar contextual representations tend to follow similar expert activation patterns. This confirms that CoVFT leverages the contextual vector as a central bridge between multimodal context and the visual encoding process. By routing context-similar samples through similar expert combinations, CoVFT effectively mitigates visual preference conflicts within the vision encoder.

\noindent \textbf{Data scalability.} We further investigate the scalability of CoVFT by conducting ablations on the amount of training data, as shown in Fig.~\ref{fig:diff_data}. Across all data scales and task categories, CoVFT consistently outperforms the baseline, demonstrating its ability to capture fine-grained contextual dependencies and maintain stable performance under varying data volumes. Notably, CoVFT surpasses the full-data LLaVA-1.5-7B baseline when trained on 75\% of the data, highlighting its strong data efficiency that is especially valuable in real-world multimodal scenarios where high-quality annotated data is limited or expensive to acquire.

A closer examination of different task groups reveals distinct scaling behaviors. For general tasks, the performance gains of CoVFT gradually saturate as data volume increases, suggesting that visual preference conflicts in these tasks are relatively mild and can be partially mitigated through large-scale data exposure. For Knowledge and OCR tasks, CoVFT's improvements increase steadily with more data, indicating robust generalization across diverse visual–linguistic contexts and a continued reduction of context-induced conflicts with richer multimodal supervision. For vision-centric tasks, CoVFT substantially enhances the capacity of the vision encoder: while the \emph{Freeze} and \emph{Full ft.} begin to plateau after 50\% of the data, CoVFT continues to deliver consistent and significant gains. These results confirm the effectiveness of context-aware visual fine-tuning, particularly in scenarios that demand strong visual reasoning and high-level cross-modal alignment.

\noindent \textbf{Generalizability.} To assess the generalizability of CoVFT across MLLMs with diverse modules and default VFT strategies, we replace the CLIP \cite{clip} encoder in LLaVA-1.5 with SigLiP \cite{siglip} and DINOv3 \cite{dinov3}, both adopted in recent MLLMs \cite{deepseekvl, cambrian}, while keeping all other settings identical. As shown in Table~\ref{model_generalize}, CoVFT consistently improves over the corresponding Freeze baselines on most benchmarks, demonstrating robustness to variations in visual representation quality and pre-training objectives. Applying CoVFT to InternVL 2.0 \cite{internvl} further yields gains across multiple benchmarks, indicating that context-aware visual adaptation generalizes beyond the LLaVA-style pipeline. Overall, these results show that CoVFT remains broadly effective across heterogeneous MLLM architectures.

\section{Conclusion}

We introduce the Context-aware Visual Fine-tuning (CoVFT) framework for fine-tuning the vision encoder within MLLMs. Guided by a systematic benchmark of existing VFT methods and an analysis of visual preference conflicts, CoVFT integrates two components: the Contextual Vector Extraction (CVE) module, which aggregates multimodal signals into a compact contextual representation, and the Contextual Mixture-of-Experts (CoMoE) module, which enables context-aware routing within the vision encoder. Experiments demonstrate that CoVFT achieves state-of-the-art performance with superior stability over existing VFT methods, while further analysis shows that CoVFT remains consistent under varying data scales and MLLM architectures.

\noindent \textbf{Limitations and Future Works.} 
While CoVFT demonstrates advantages, several limitations remain. Our evaluation focuses on models up to 13B parameters and million-scale training data due to computational constraints. Although the observed scalability suggests promising potential, exploring the behavior and upper bound of visual fine-tuning in larger MLLMs is an important direction. In addition, reinforcement learning has recently shown strong potential for post-training MLLMs, while vision encoder optimization under such settings still suffer from instability \cite{rl_tune}. Developing stable visual encoder training strategies for RL-based MLLMs is therefore a promising direction for future work, and may contribute to more unified training pipelines for multimodal systems.

\section*{Acknowledgment} 
This work is partly supported by the National Key Research and Development Plan (2024YFB3309302), the National Natural Science
Foundation of China (82441024), the Research Program of State Key Laboratory of Complex and Critical Software Environment, and the Fundamental Research Funds for the Central Universities.

The authors sincerely thank Hebeizi Li and Xiefan Guo for insightful discussions on the framework design, Zhenghao Chen for assistance with the training framework, and the anonymous reviewers for their constructive comments.

{
    \small
    \bibliographystyle{ieeenat_fullname}
    \bibliography{main}
}

\appendix
\clearpage

\twocolumn[
\begin{center}
    \Large\bfseries Supplementary Material for \emph{CoVFT: Context-aware Visual Fine-tuning for Multimodal Large Language Models}
\end{center}
\vspace{0.5em}
]

\noindent This supplementary document provides additional experimental results and extended analyses that complement the main paper. The content is organized as follows:  
\begin{itemize}[leftmargin=*]
    \item \textbf{Section~A} presents extended ablation studies of CoVFT, including additional design choices and hyper-parameters.  
    \item \textbf{Section~B} presents extended analyzes on the gradient conflict, context-aware counterparts, learnable parameters and computational overhead.
    \item \textbf{Section~C} provides visualizations of contextual vectors and multimodal samples.  
    \item \textbf{Section~D} reports the complete benchmark results for all evaluated tasks, extending the averaged results shown in the main paper.
\end{itemize}

\section{Extended Ablation Studies}
\label{sec:ext_ablation}

\noindent In this section, we provide additional analyses of key design choices in CoVFT, focusing on the placement of CoMoE layers and the number of experts per CoMoE block.

\paragraph{Placement of CoMoE layers.}
We evaluate the impact of different CoMoE placements within the vision encoder in Table~\ref{tab:comoe_layer}. Since only the first 23 ViT blocks participate in the LLaVA-1.5 forward pass \cite{llava1.5}, we index them from 0 to 22 and examine four representative configurations: inserting CoMoE into \textit{all} eligible layers (0–22), into the \textit{first half} (0–10), into the \textit{second half} (11–22), and into progressively deeper subsets of layers.
From Table~\ref{tab:comoe_layer}, we observe that introducing CoMoE consistently improves performance over the baseline (\emph{N/A}), which corresponds to full fine-tuning without CoMoE, demonstrating its overall effectiveness. Furthermore, deeper CoMoE placement generally yields larger gains under comparable layer budgets. This trend aligns with the observation in Fig.~1(c), where visual preference conflicts become increasingly pronounced in deeper layers of the vision encoder. We additionally evaluate several reduced-layer variants and find that placing CoMoE in layers 11–22 delivers the most consistent improvements across benchmarks. Consequently, we adopt this configuration as the default setting for CoVFT.

\begin{table}[t]\small  
    \centering 
    \caption{Ablation on the placement of CoMoE layers within the vision encoder. 
    ``Start'' and ``End'' denote the layer range where CoMoE is inserted. 
    ``Num.'' is the total number of CoMoE layers applied. 
    The baseline (N/A) corresponds to full fine-tuning without CoMoE.}
    \setlength{\tabcolsep}{3.0pt}
    \begin{tabular}{c c c c c c c} 	
        \toprule
        Start & End & Num. &General & {Know.\&OCR} & Vision & Mean (\%)\\
        \midrule
         & N/A &  & 66.69 & 61.29 & 52.17 & 59.29 \\
        \midrule
        0 & 22 & 23 & 66.72 & 61.43 & 54.32 & 60.23 \\
        0 & 10 & 11 & 66.75 & 61.43 & 52.13 & 59.33 \\
        11 & 22 & 12 & \cellcolor{gray!20}\textbf{67.04} & \cellcolor{gray!20}\textbf{61.93} & \cellcolor{gray!20}\textbf{55.81} & \cellcolor{gray!20}\textbf{61.08} \\
        15 & 22 & 8 & 66.96 & 62.10 & 52.87 & 59.87 \\
        19 & 22 & 4 & 66.78 & 61.70 & 53.50 & 59.98 \\
        \bottomrule 
    \end{tabular}
    \label{tab:comoe_layer}
\end{table}

\begin{table}[t]\small  
    \centering 
    \caption{Ablation on the number of experts in each CoMoE block. 
    }
    \setlength{\tabcolsep}{3.0pt}
    \begin{tabular}{c c c c c} 	
        \toprule
        Num. &General & {Know.\&OCR} & Vision & Mean (\%)\\
        \midrule
        2 & 66.24 & 61.63 & 55.16 & 60.47 \\
        4 & \cellcolor{gray!20}\textbf{67.04} & \cellcolor{gray!20}\textbf{61.93} & 55.81 & \cellcolor{gray!20}\textbf{61.08} \\
        8 & 66.12 & 61.73 & \cellcolor{gray!20}\textbf{56.03} & 60.82 \\
        \bottomrule 
    \end{tabular}
    \label{tab:expert_num}
\end{table}
\paragraph{Number of experts.} Table~\ref{tab:expert_num} reports an ablation on the number of experts in each CoMoE block. Using four experts yields the best mean performance among the tested configurations. We note that the optimal number of experts is related to the scale and diversity of training data \cite{moe_scaling}. The hyperparameters in Table~\ref{tab:expert_num} are determined under an instruction-tuning set of roughly 665K samples. Based on existing MoE scaling observations \cite{moe_scaling,moe_scaling_2}, we expect that larger-scale training may benefit from increasing the number of experts, and adjusting expert width according to empirical scaling rules may further improve performance. 

To further investigate the optimal MoE capacity under different levels of training data diversity, we partition the LLaVA-1.5 instruction data into 15 task types using Qwen-Plus, based on perceptual focus, semantic granularity, and reasoning requirements, and then ablate the number of experts. The results are shown in Table~\ref{tab:moe_capacity}. As task diversity increases (3$\rightarrow$15), the optimal number of experts also increases (2$\rightarrow$4$\rightarrow$8), indicating that the optimal MoE capacity scales with data diversity. This trend suggests that CoMoE allocates more specialized expert subspaces to accommodate heterogeneous visual preferences induced by increasingly diverse multimodal contexts.

\begin{table}[h]
    \centering
    \footnotesize 
    \caption{Vision-centric task performance under different levels of task diversity and numbers of experts. \textbf{Bold} indicates the best configuration for each diversity level.}
    \setlength{\tabcolsep}{4.5pt} 
    \begin{tabular}{@{}c|cccccc@{}}
        \toprule
        \textbf{\# Experts} &\textbf{\# Task types$\rightarrow$}  & {3} & {6} & {9} & {12} & {15}  \\
        \midrule
        Full ft. &  & 45.49 & 47.66 & 49.01 & 51.79 & 52.17 \\
         2 & & \cellcolor{gray!20}\textbf{48.09} & \cellcolor{gray!20}\textbf{49.55} & 51.09 & 53.14 & 55.16 \\
         4 & & 47.36 & 48.71 & \cellcolor{gray!20}\textbf{52.27} & \cellcolor{gray!20}\textbf{54.72} & 55.81 \\
        8 & & 46.94 & 48.35 & 52.01 & 54.50 & \cellcolor{gray!20}\textbf{56.03} \\ 
        \bottomrule
    \end{tabular}
    \label{tab:moe_capacity}
\end{table}

\section{Extended Analysis}
\label{sec:computational_overhead}

\begin{figure}[h]
  \centering
  \includegraphics[width=82mm]{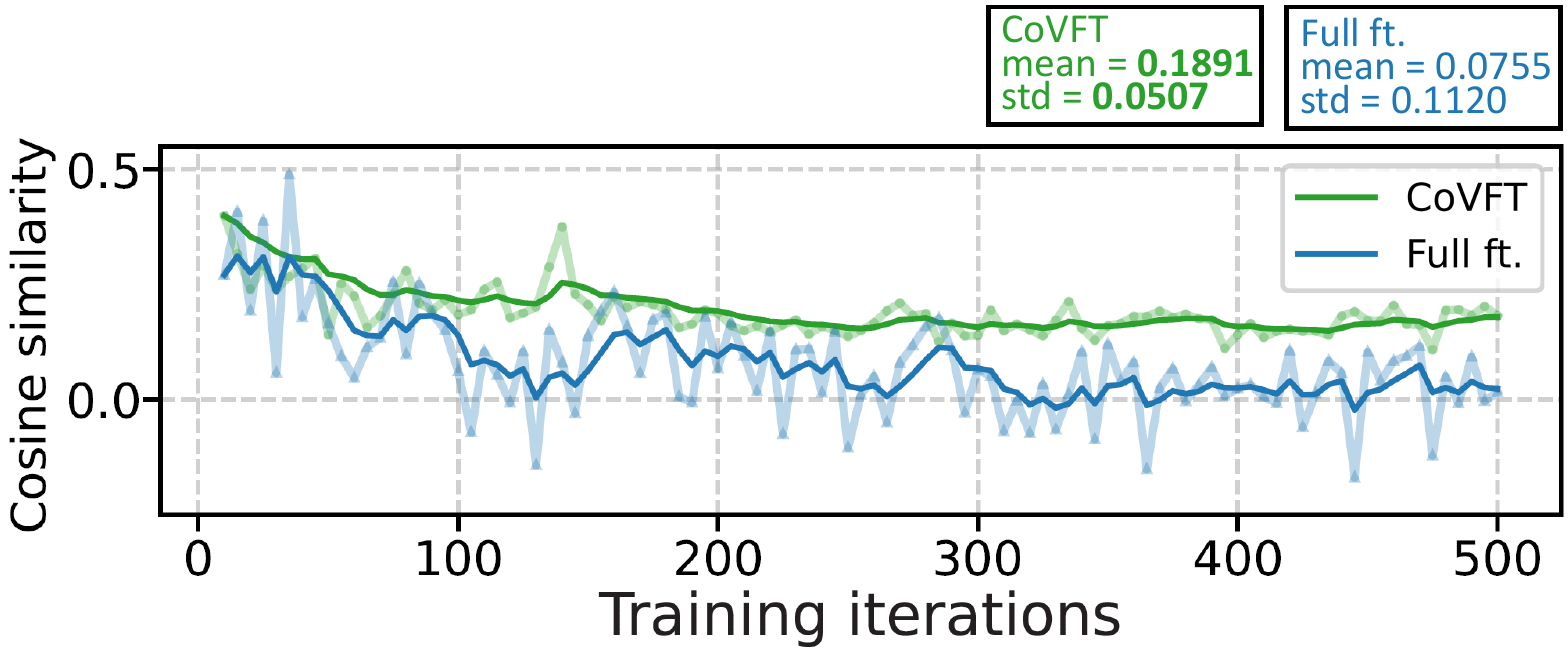}
  \caption{Cosine similarity between each gradient update and the dominant gradient direction during training under the standard instruction-tuning setting.}
  \label{gradient}
\end{figure}

\textbf{Gradient conflict.} In Fig.~1(a) of the main body, we analyze visual preference conflict from the perspective of parameter distance. Here, we further provide an analysis from the gradient perspective. Specifically, we compute the cosine similarity between each gradient update and the dominant gradient direction during training under the standard mixed-task setting. The results are shown in Fig.~\ref{gradient}. Full fine-tuning exhibits frequent near-orthogonal or even opposing gradients, indicating update conflicts across heterogeneous instructions. In contrast, CoVFT substantially increases the mean gradient similarity (0.076$\rightarrow$0.189) while reducing the standard deviation (0.112$\rightarrow$0.051), suggesting that it improves gradient alignment and stabilizes optimization.

\noindent \textbf{Comparison with context-aware counterparts.} To further compare with existing context-aware counterparts, we re-implement QA-ViT~\cite{qavit} and Q-MoE~\cite{qmoe} within LLaVA-1.5. CoVFT achieves \textbf{61.08}\% mean accuracy, outperforming QA-ViT (59.12\%) and Q-MoE (59.29\%) across 12 benchmarks, demonstrating the advantage of our method for context-aware visual adaptation in MLLMs.

\noindent \textbf{Parameter-matched comparison.} We compare different methods under matched trainable parameter budgets. With 4 CoMoE layers, CoVFT matches LoRA ($r{=}512$) in trainable parameters (6.85B vs. 6.86B) while achieving better overall performance (59.98 vs. 59.20), especially on vision-centric tasks (53.50 vs. 52.67). When increasing the number of CoMoE layers to 12, CoVFT continues to scale effectively, improving overall performance from 59.98 to 61.08 and vision-centric performance from 53.50 to 55.81. In contrast, existing PETL methods are more difficult to scale in this manner, highlighting the superior scalability and effectiveness of CoVFT under similar parameter budgets.

\begin{table}[h]\footnotesize  
    \centering 
    \setlength{\tabcolsep}{4.5pt}
    \caption{Comparison of diverse VFT methods under different trainable parameters. $N$ denotes the number of CoMoE layers.}
    \begin{tabular}{l c c c c c} 	
        \toprule
        Method & G. & K. & V. & Mean (\%) & Param. (B)\\
        \midrule
        Freeze & 66.23 & 61.20 & 51.71 & 58.93 & 6.76 \\
        LoRA ($r$=8) &  65.93 & 60.86 & 52.45 & 59.04 & 6.76 \\
        LoRA ($r$=512) & 66.09 & 60.90 & 52.67 & 59.20 & 6.86 \\
        CoVFT ($N$=4)& \cellcolor{gray!20}\textbf{66.78} & \cellcolor{gray!20}\textbf{61.70} & \cellcolor{gray!20}\textbf{53.50} & \cellcolor{gray!20}\textbf{59.98} & 6.85 \\
        \midrule
        Full fine-tuning & 66.69 & 61.29 & 52.17 & 59.29 & 7.06 \\
        CoVFT ($N$=12) & \cellcolor{gray!20}\textbf{67.04} & \cellcolor{gray!20}\textbf{61.93} & \cellcolor{gray!20}\textbf{55.81} & \cellcolor{gray!20}\textbf{61.08} & 7.24 \\
        \bottomrule 
    \end{tabular}
\end{table}

\paragraph{Computational overhead.} We analyze the computational overhead of CoVFT and compare it with two widely used VFT baselines: \emph{Freeze} and \emph{Full fine-tuning}. All pre-training and instruction-tuning experiments are conducted on 8$\times$NVIDIA H100 GPUs, with batch sizes identical to those used in LLaVA-1.5. For inference, we evaluate each method on a single NVIDIA H100 GPU with a batch size of 1. 

The computational cost comparison among \emph{Freeze}, \emph{Full FT}, and \emph{CoVFT} is reported in Table~\ref{tab:computing}. Compared with the freeze setting and full fine-tuning, CoVFT introduces 18 minutes in total training time (approximately 3.8\% overhead relative to Full FT), adds $\sim$10ms latency during inference, and incurs a moderate increase in peak GPU memory usage (13.5\% over Full FT). The frozen BERT encoder incurs around 1.14ms inference latency per sample. Importantly, this slight computational and memory overhead yields substantial accuracy gains across 12 multimodal benchmarks and delivers markedly improved stability, offering a significantly better trade-off among performance, robustness, and efficiency. This demonstrates that context-aware visual adaptation is both practical and cost-effective for modern MLLMs. Moreover, CoVFT remains fully compatible with existing MoE-parallelization and expert sharding techniques~\cite{moe_parallelism}, which can further reduce training, inference, and memory costs, suggesting that its efficiency can be improved even further in future deployments.

\begin{table*}[h]\small  
    \centering 
    \caption{Comparison of computational overhead across different VFT strategies. 
    Training time is reported in hours and minutes; inference time is averaged per sample.}
    \setlength{\tabcolsep}{3.5pt}
    \begin{tabular}{l c c c c c} 	
        \toprule
        Method & Pre-training & Instruction-tuning& Total & Inference & Peak GPU memory\\
        \midrule
        Freeze & 2h 38m & 4h 28m & 7h 6m & 77.04ms & 41,352MB \\
        Full ft. & 2h 38m & 5h 14m & 7h 52m & 77.04ms & 44,036MB \\
        CoVFT & 2h 38m & 5h 32m & 8h 10m & 88.17ms & 49,974MB\\
        \bottomrule 
    \end{tabular}
    \label{tab:computing}
\end{table*}

\begin{figure*}[t]
  \centering 
  \includegraphics[width=174mm]{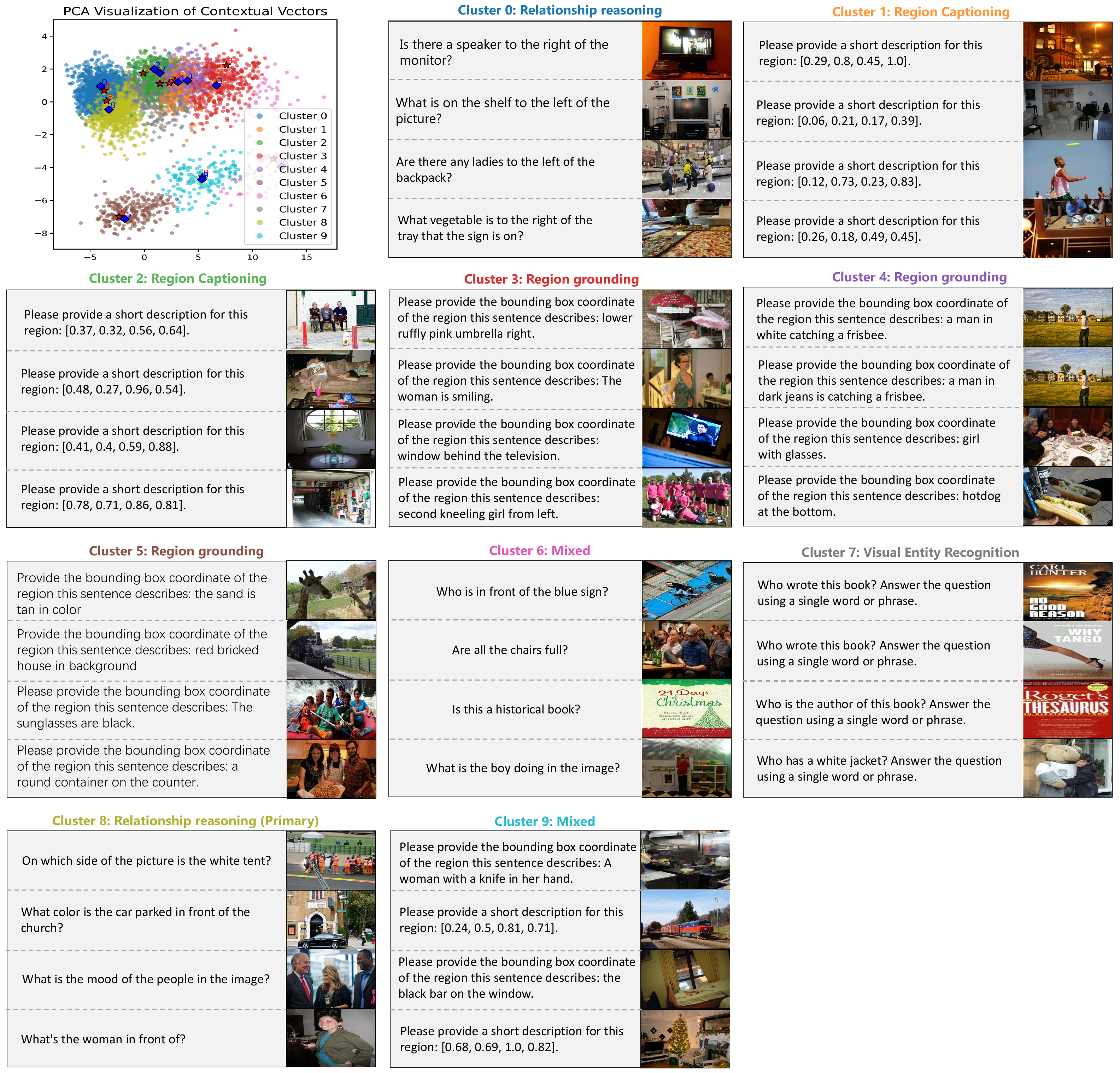}
  \caption{PCA~\cite{pca} visualization of 5,000 contextual vectors extracted from the CVE module. Vectors are grouped using \emph{k}-means~\cite{kmeans}, and for each cluster, four nearest samples to the cluster centroid are shown to illustrate representative task patterns.
} 
  \label{fig:analysis_fig_supp}
\end{figure*}

\section{Visualizations}
\label{sec:visualizations}

In the main paper (Fig.~3(a)), we visualize the contextual vectors extracted from the CVE module, showing clear semantic grouping aligned with distinct visual preference patterns. Here, we extend this analysis by presenting the full clustering results in Fig.~\ref{fig:analysis_fig_supp}, together with representative samples from each cluster.

\paragraph{Visualization procedure.}
We randomly sample 5,000 multimodal instruction–image pairs from the LLaVA-665K instruction-tuning dataset and extract their contextual vectors from the penultimate vision encoder layer (i.e., the features actually fed to the LLM). These vectors are grouped using \emph{k}-means~\cite{kmeans} clustering, projected into two dimensions using PCA~\cite{pca} for visualization, and we select the four samples nearest to each cluster centroid to illustrate the dominant characteristics of each cluster.

\paragraph{Observed cluster patterns.}

As shown in Fig.~\ref{fig:analysis_fig_supp}, the contextual vectors exhibit a clear task-dependent organization: clusters naturally align with distinct types of multimodal reasoning despite the absence of explicit task labels. For example, Clusters 0 and 8 correspond to \emph{relationship reasoning} involving structural or relational understanding across objects; Clusters 1 and 2 capture \emph{region captioning}, focusing on localized box-level descriptions; and Clusters 3, 4, and 5 represent \emph{region grounding}, mapping textual phrases to spatial coordinates. Other clusters display more heterogeneous behaviors: Cluster 7 leans toward \emph{visual entity recognition}, including OCR and attribute-based identification; Cluster 9 frequently mixes grounding and captioning; while Cluster 6 spans diverse queries without a dominant pattern. These emergent structures indicate that CVE effectively organizes the multimodal space into semantically meaningful subregions, capturing latent task semantics purely from contextual signals. Building upon these structured contextual embeddings, CoMoE further routes samples to expert pathways that better align the vision encoder’s updates with multimodal instruction-following objectives while reducing conflicting optimization signals.

\paragraph{Discussion.}
While CVE provides an effective mechanism for capturing contextual signals, the clustering results suggest room for further refinement. Some clusters (e.g., Cluster~6) exhibit coarse or mixed semantics, suggesting that contextual modeling could be made more discriminative. Conversely, overly fine-grained contextual partitioning may also hinder expert-level knowledge sharing. Balancing fine-grained modeling of visual preferences with the need for shared representation learning, as well as understanding how longer and more complex textual contexts shape these clusters, offers promising directions for future research.

\section{Full Results}
\label{sec:full_results}

For reasons of space, the main paper reports the averaged results for several groups of benchmarks. In this section, we provide the complete per-task results corresponding to all experiments in the main text\footnote{Following \cite{cambrian}, we divide the MME Perception score by 20 to have the same scale as other benchmarks before averaging.}. Specifically, Table~\ref{per_task_ablation} presents the full results for the design choices of the contextual vector and routing strategy (Table~2 in the main paper). Table~\ref{per_task_diff_data} reports the per-task results under different data scales (Figure~4 in the main paper). Table~\ref{per_task_comoe_place} and Table~\ref{per_task_diff_expert} provide the detailed results for the ablations on the placement of CoMoE layers and the number of experts, respectively, corresponding to Tables~\ref{tab:comoe_layer} and~\ref{tab:expert_num} of the supplementary material.

\begin{table*}[t]\footnotesize
	\centering 
  \setlength{\tabcolsep}{2.7pt}
  \caption{Per task results by various ablation settings in Table 2 of the main body.}
	\begin{tabular}{l c c c c c c c c c c c c c c c c} 
      \toprule     
      ~ & \multicolumn{4}{c}{\textbf{General}} & \multicolumn{3}{c}{\textbf{Knowledge \& OCR}} & \multicolumn{5}{c}{ \textbf{Vision-centric}} & \multicolumn{4}{c}{\textbf{Mean (\%)}} \\
      \cmidrule(lr{0pt}){2-5} \cmidrule(lr{0pt}){6-8} \cmidrule(lr{0pt}){9-13} \cmidrule(lr{0pt}){14-17}
      Ablation & MME$^{\text P}$ & MMB$^\text{en}$ & MMB$^\text{cn}$ & GQA & SQA$^\text{I}$ & AI2D & TextVQA & MMVP & RWQA & COCO & ADE & Omni3D & \textbf{G.} & \textbf{K.} & \textbf{V.} & \textbf{Avg.} \\ 
      \midrule
      \multicolumn{5}{l}{\textcolor{gray}{\emph{w.r.t.} contextual vector}}\\
      \makecell[l]{N/A} & {1510.2} & 67.44 & 59.88 & {63.92} & 67.67 & 56.51 & {59.70} & 27.33 & 55.82 & {65.71} & 51.97 & 60.00 & {66.69} & 61.29 & 52.17 & 59.29\\
      \makecell[l]{Image-only} & 1492.2 & 68.56 & 61.17 & 62.04 & 68.96 & 56.80 & 59.32 & 26.00 & 56.99 & 66.71 & 55.13 & 61.00 & 66.60 & 61.69 & 53.17 & 59.77\\
      \makecell[l]{Text-only} & 1505.6 & 69.07 & 60.22 & 62.77 & 68.72 & 57.35 & 59.52 & 30.66 & 56.99 & 68.82 & 57.35 & 59.83 & 66.84 & 61.86 & 54.73 & 60.55\\
      \makecell[l]{Concat$[\rm I,T]$} & 1498.6 & 68.90 & 60.48 & 62.82 & 69.16 & 56.67 & 59.54 & 31.33 & 55.82 & 68.70 & 55.29 & 61.67 & 66.78 & 61.79 & 54.56 & 60.44\\
      \makecell[l]{CVE} & {1525.2} & {68.13} & {60.40} & 63.37 & {69.51} & {56.64} & {59.64} & {36.67} & {57.52} & {66.96} & {56.08} & {61.83} & {67.04} & {61.93} & {55.81} & {61.08}\\
      \midrule
      \multicolumn{5}{l}{\textcolor{gray}{\emph{w.r.t.} routing strategy}}\\
      \makecell[l]{Uniform} & {1478.8} & {67.61} & {59.28} & 63.89 & {69.26} & {56.80} & {59.20} & {28.67} & {57.65} & {66.21} & {51.03} & {61.67} & 66.18 & 61.75 & 53.05 & 59.60\\
      \makecell[l]{Sparse@2} & {1478.8} & {69.07} & {60.57} & 62.91 & {69.91} & {56.67} & {60.15} & {30.00} & {56.21} & {67.20} & {51.82} & {62.75} & 66.63 & 61.78 & 53.60 & 60.10\\
      \makecell[l]{Dense} & {1525.2} & {68.13} & {60.40} & 63.37 & {69.51} & {56.64} & {59.64} & {36.67} & {57.52} & {66.96} & {56.08} & {61.83} & {67.04} & {61.93} & {55.81} & {61.08}\\
      \bottomrule 
    \end{tabular}
  
  \label{per_task_ablation}
\end{table*}

\begin{table*}[t]\footnotesize
	\centering 
  \setlength{\tabcolsep}{2.7pt}
  \caption{
    Per task results by various ablation settings in Figure 4 of the main body.
  }
	\begin{tabular}{l c c c c c c c c c c c c c c c c} 
      \toprule     
      ~ & \multicolumn{4}{c}{\textbf{General}} & \multicolumn{3}{c}{\textbf{Knowledge \& OCR}} & \multicolumn{5}{c}{ \textbf{Vision-centric}} & \multicolumn{4}{c}{\textbf{Mean (\%)}} \\
      \cmidrule(lr{0pt}){2-5} \cmidrule(lr{0pt}){6-8} \cmidrule(lr{0pt}){9-13} \cmidrule(lr{0pt}){14-17}
      Method & MME$^{\text P}$ & MMB$^\text{en}$ & MMB$^\text{cn}$ & GQA & SQA$^\text{I}$ & AI2D & TextVQA & MMVP & RWQA & COCO & ADE & Omni3D & \textbf{G.} & \textbf{K.} & \textbf{V.} & \textbf{Avg.} \\ 
      \midrule
      \multicolumn{5}{l}{\textcolor{gray}{100\% data}}\\
      \makecell[l]{Freeze} & 1473.7 & {67.87} & {60.30} & 63.07 & {69.31} & 55.76 & 58.53 & 28.00 & {56.73} & 63.73 & 49.61 & 60.50 & 66.23 & 61.20 & 51.71 & 58.93\\
      \makecell[l]{Full fine-tuning} & {1510.2} & 67.44 & 59.88 & {63.92} & 67.67 & 56.51 & {59.70} & 27.33 & 55.82 & {65.71} & 51.97 & 60.00 & {66.69} & 61.29 & 52.17 & 59.29\\
      \makecell[l]{CoVFT} & {1525.2} & {68.13} & {60.40} & 63.37 & {69.51} & {56.64} & {59.64} & {36.67} & {57.52} & {66.96} & {56.08} & {61.83} & {67.04} & {61.93} & {55.81} & {61.08}\\
      \midrule
      \multicolumn{5}{l}{\textcolor{gray}{50\% data}}\\
      \makecell[l]{Freeze} & 1446.0 & 65.55 & 57.47 & 60.84 & 68.96 & 55.63 & 57.01 & 28.00 & 53.46 & 64.72 & 49.45 & 59.42 & 64.04 & 60.53 & 51.01 & 57.73\\
      \makecell[l]{Full fine-tuning} & 1445.2 & 67.18 & 56.87 & 62.23 & 66.09 & 57.35 & 57.40 & 29.33 & 53.59 & 65.96 & 52.13 & 58.00 & 64.63 & 60.28 & 51.80 & 58.20\\
      \makecell[l]{CoVFT} & {1492.0} & {67.7} & {58.08} & 61.25 & {68.32} & {56.83} & {58.07} & {26.00} & {55.29} & {66.96} & {49.92} & {58.17} & {65.41} & {61.07} & {51.27} & {58.43}\\
      \midrule
      \multicolumn{5}{l}{\textcolor{gray}{25\% data}}\\
      \makecell[l]{Freeze} & 1463.4 & 62.89 & 55.07 & 58.67 & 67.82 & 54.92 & 55.54 & 19.33 & 51.63 & 57.64 & 47.39 & 60.25 & 62.45 & 59.43 & 47.25 & 55.36\\
      \makecell[l]{Full fine-tuning} & 1397.6 & 63.32 & 53.18 & 60.43 & 67.53 & 54.76 & 56.24 & 24.67 & 50.72 & 57.76 & 47.87 & 62.92 & 61.70 & 59.51 & 48.79 & 55.77\\
      \makecell[l]{CoVFT} & {1467.0} & {64.35} & {54.90} & 59.91 & {67.13} & {55.60} & {56.93} & {23.33} & {53.73} & {57.27} & {47.39} & {59.83} & {63.13} & {59.89} & {48.31} & {56.14}\\
      \midrule
      \multicolumn{5}{l}{\textcolor{gray}{12.5\% data}}\\
      \makecell[l]{Freeze} & 1430.6 & 59.19 & 53.09 & 56.11 & 66.68 & 54.60 & 53.38 & 15.33 & 47.97 & 55.53 & 49.29 & 58.17 & 59.98 & 58.55 & 45.26 & 53.49\\
      \makecell[l]{Full fine-tuning} & 1408.2 & 57.82 & 50.52 & 57.93 & 66.63 & 53.79 & 54.50 & 22.00 & 50.46 & 54.78 & 46.45 & 58.50 & 59.17 & 58.31 & 46.44 & 53.65\\
      \makecell[l]{CoVFT} & {1382.2} & {59.11} & {50.69} & 57.98 & {66.34} & {53.95} & {55.37} & {21.33} & {49.93} & {59.38} & {48.18} & {59.50} & {59.22} & {58.55} & {47.66} & {54.24}\\
      \bottomrule 
    \end{tabular}
  
  \label{per_task_diff_data}
\end{table*}

\begin{table*}[t]\footnotesize
	\centering 
  \setlength{\tabcolsep}{2.5pt}
  \caption{
    Per task results by various ablation settings in Table~\ref{tab:comoe_layer} of the supplementary material.
  }
	\begin{tabular}{l c c c c c c c c c c c c c c c c c c} 
      \toprule     
      ~ & ~ & ~ & \multicolumn{4}{c}{\textbf{General}} & \multicolumn{3}{c}{\textbf{Knowledge \& OCR}} & \multicolumn{5}{c}{ \textbf{Vision-centric}} & \multicolumn{4}{c}{\textbf{Mean (\%)}} \\
      \cmidrule(lr{0pt}){4-7} \cmidrule(lr{0pt}){8-10} \cmidrule(lr{0pt}){11-15} \cmidrule(lr{0pt}){16-19}
      Start & End & Num. & MME$^{\text P}$ & MMB$^\text{en}$ & MMB$^\text{cn}$ & GQA & SQA$^\text{I}$ & AI2D & TextVQA & MMVP & RWQA & COCO & ADE & Omni3D & \textbf{G.} & \textbf{K.} & \textbf{V.} & \textbf{Avg.} \\ 
      \midrule
       & N/A &   & {1510.2} & 67.44 & 59.88 & {63.92} & 67.67 & 56.51 & {59.70} & 27.33 & 55.82 & {65.71} & 51.97 & 60.00 & {66.69} & 61.29 & 52.17 & 59.29\\
       \midrule
      0 & 22 &  23 & 1508.4 & 68.08 & 60.55 & 62.84 & 67.92 & 56.93 & 59.45 & 30.67 & 55.69 & 67.20 & 55.29 & 62.75 & 66.72 & 61.43 & 54.32 & 60.23\\
      0 & 10 & 11 & 1501.4 & 68.04 & 60.65 & 63.22 & 68.62 & 56.99 & 58.68 & 27.33 & 56.86 & 62.11 & 52.34 & 62.00 & 66.75 & 61.43 & 52.13 & 59.33 \\
      11 & 22 & 12 & {1525.2} & {68.13} & {60.40} & 63.37 & {69.51} & {56.64} & {59.64} & {36.67} & {57.52} & {66.96} & {56.08} & {61.83} & {67.04} & {61.93} & {55.81} & {61.08}\\
      15 & 22 & 8  & 1518.0 & 68.21 & 60.45 & 63.29 & 69.66 & 56.77 & 59.88 & 31.33 & 56.60 & 64.57 & 54.50 & 57.33 & 66.96 & 62.10 & 52.87 & 59.87\\
      19 & 22 & 4  & 1513.2 & 68.10 & 60.33 & 63.01 & 69.36 & 56.25 & 59.50 & 29.33 & 56.60 & 66.71 & 55.13 & 59.75 & 66.78 & 61.70 & 53.50 & 59.98\\
      \bottomrule 
    \end{tabular}
  
  \label{per_task_comoe_place}
\end{table*}

\begin{table*}[t]\footnotesize
	\centering 
  \setlength{\tabcolsep}{3.7pt}
  \caption{
    Per task results by various ablation settings in Table~\ref{tab:expert_num} of the supplementary material.
  }
	\begin{tabular}{l c c c c c c c c c c c c c c c c} 
      \toprule     
      ~ & \multicolumn{4}{c}{\textbf{General}} & \multicolumn{3}{c}{\textbf{Knowledge \& OCR}} & \multicolumn{5}{c}{ \textbf{Vision-centric}} & \multicolumn{4}{c}{\textbf{Mean (\%)}} \\
      \cmidrule(lr{0pt}){2-5} \cmidrule(lr{0pt}){6-8} \cmidrule(lr{0pt}){9-13} \cmidrule(lr{0pt}){14-17}
      Num. & MME$^{\text P}$ & MMB$^\text{en}$ & MMB$^\text{cn}$ & GQA & SQA$^\text{I}$ & AI2D & TextVQA & MMVP & RWQA & COCO & ADE & Omni3D & \textbf{G.} & \textbf{K.} & \textbf{V.} & \textbf{Avg.} \\ 
      \midrule
      \makecell[l]{2} & 1478.6 & 67.61 & 60.91 & 62.51 & 69.06 & 56.25 & 59.59 & 32.00 & 58.04 & 68.57 & 55.77 & 61.42 & 66.24 & 61.63 & 55.16 & 60.47 \\
      \makecell[l]{4} & {1525.2} & {68.13} & {60.40} & 63.37 & {69.51} & {56.64} & {59.64} & {36.67} & {57.52} & {66.96} & {56.08} & {61.83} & {67.04} & {61.93} & {55.81} & {61.08}\\
      \makecell[l]{8} & 1476.2 & 67.61 & 60.48 & 62.57 & 68.62 & 56.99 & 59.58 & 30.67 & 56.73 & 70.68 & 58.14 & 63.92 & 66.12 & 61.73 & 56.03 & 60.82\\
      \bottomrule 
    \end{tabular}
  
  \label{per_task_diff_expert}
\end{table*}


\end{document}